\documentclass[sigconf]{acmart}
\usepackage{multirow}
\AtBeginDocument{%
  }

\setcopyright{acmcopyright}
\copyrightyear{2023}
\acmYear{2023}
\acmDOI{XXXXXXX.XXXXXXX}

\acmConference[MM '23]{Proceedings of the 31th ACM International Conference on Multimedia}{June 03--05,
  2018}{Woodstock, NY}
\acmBooktitle{Proceedings of the 31th ACM International Conference on Multimedia (MM '23),
  June 03--05, 2018, Woodstock, NY}
\acmPrice{15.00}
\acmISBN{978-1-4503-XXXX-X/18/06}

\acmSubmissionID{2069}

\begin{document}

\title{PEARL: Preprocessing Enhanced Adversarial Robust Learning of Image Deraining for Semantic Segmentation}


\author{Xianghao Jiao}
\affiliation{%
  \institution{Dalian University of Technology}
  \city{Dalian}
  \state{Liaoning}
  \country{China}}
\email{jxh@mail.dlut.edu.cn}

\author{Yaohua Liu}
\affiliation{%
	\institution{Dalian University of Technology}
	\city{Dalian}
	\state{Liaoning}
	\country{China}}
\email{liuyaohua_918@163.com}

\author{Xinjia Gao}
\affiliation{%
	\institution{Dalian University of Technology}
	\city{Dalian}
	\state{Liaoning}
	\country{China}}
\email{jiaxinn.gao@outlook.com}

\author{Xinyuan Chu}
\affiliation{%
\institution{Dalian University of Technology}
 \city{Dalian}
 \state{Liaoning}
 \country{China}}
\email{chuxinyuan_dut@outlook.com}

\author{Xin Fan}
\affiliation{%
	\institution{Dalian University of Technology}
	\city{Dalian}
	\state{Liaoning}
	\country{China}}
\email{xin.fan@dlut.edu.cn}

\author{Risheng Liu}
\affiliation{%
	\institution{Dalian University of Technology}
	\city{Dalian}
	\state{Liaoning}
	\country{China}}
\email{rsliu@dlut.edu.cn}

%

\renewcommand{\shortauthors}{Trovato et al.}

\begin{abstract}
	
In light of the significant progress made in the development and application of semantic segmentation tasks, there has been increasing attention towards improving the robustness of segmentation models against natural degradation factors (e.g., rain streaks) or artificially attack factors (e.g., adversarial attack). Whereas, most existing methods are designed to address a single degradation factor and are tailored to specific application scenarios. In this work, we present the first attempt to improve the robustness of semantic segmentation tasks by simultaneously handling different types of degradation factors. Specifically, we introduce the Preprocessing Enhanced Adversarial Robust Learning (PEARL) framework based on the analysis of our proposed Naive Adversarial Training (NAT) framework. Our approach effectively handles both rain streaks and adversarial perturbation by transferring the robustness of the segmentation model to the image derain model. Furthermore, as opposed to the commonly used Negative Adversarial Attack (NAA), we design the Auxiliary Mirror Attack (AMA) to introduce positive information prior to the training of the PEARL framework, which improves defense capability and segmentation performance. Our extensive experiments and ablation studies based on different derain methods and segmentation models have demonstrated the significant performance improvement of PEARL with AMA in defense against various adversarial attacks and rain streaks while maintaining high generalization performance across different datasets.

\end{abstract}




\received{20 February 2007}
\received[revised]{12 March 2009}
\received[accepted]{5 June 2009}

\begin{teaserfigure}
	\centering
	\begin{tabular}{c}
		\includegraphics[height=4.7cm,width=17.51cm,trim=10 180 10. 100,clip]{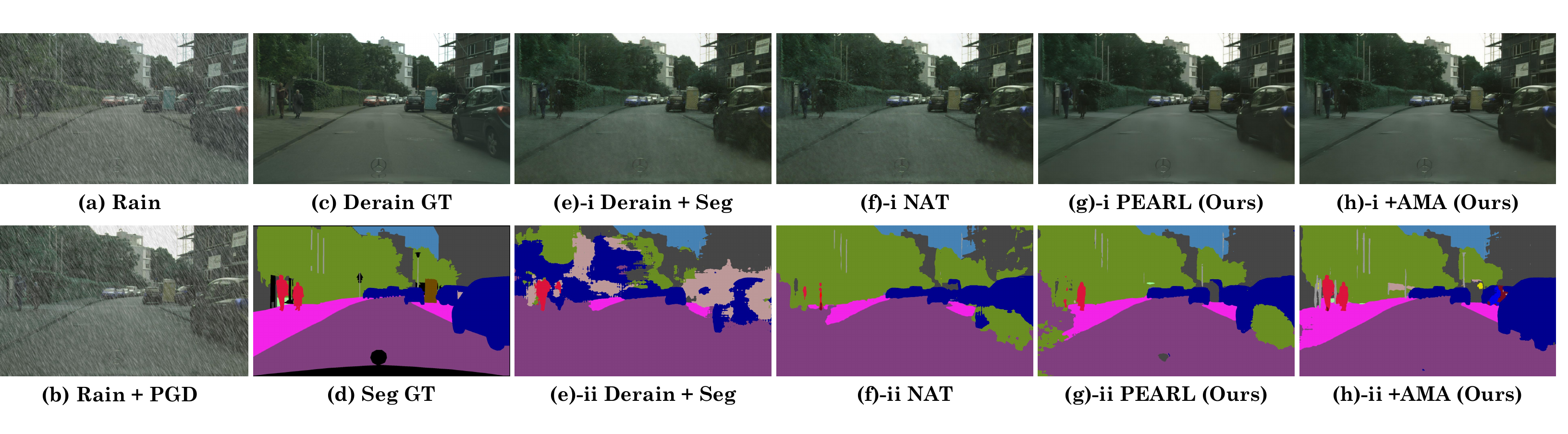} \\
	\end{tabular}
	\vspace{-0.5cm} 
	\caption{The visualization results of image deraining and semantic segmentation tasks among the baseline (Derain + Seg) and our proposed NAT framework, PEARL framework and PEARL with AMA generator (denoted as +AMA) with the influence of both degradation factors, i.e., rain streaks and PGD attacks. It can be obviously seen that our proposed framework obtains derained images with higher quality which also leads to more accurate segmentation labels.}
	\vspace{0.5cm} 
	\label{fig:firstfig}
\end{teaserfigure}
\maketitle


\section{Introduction}

Semantic segmentation~\cite{li2017instance,zhang2018context,yang2018denseaspp}, which is regarded as pixel-wise dense classification tasks to clarify each part of an image based on what category or object it belongs to, have achieved great advances with the development of deep learning networks and high-quality collected data. Meanwhile, the significant performance improvement of segmentation models also boost its application to satisfy various real-world demands~\cite{zhou2020aglnet,liu2020importance,sagar2021semantic,gonzalez2020nextmed}, e.g., self-driving systems, virtual reality, etc. Whereas, as various adversarial attacks and variations have been explored, the vulnerability of deep neural networks to these adversarial examples~\cite{wang2022generating,zhang2022towards,kim2022defending} also have attracted much attention. By introducing imperceptible adversarial perturbation to the input of semantic segmentation model, the segmentation results~\cite{xiao2018characterizing,arnab2018robustness,xie2017adversarial,hendrik2017universal} could be badly corrupted and results in serious safety issues. Moreover, since most existing methods~\cite{chen2018encoder,yu2020context} developed their approaches relying on assumption of degradation-free scenarios~\cite{cordts2016cityscapes}, the performance of segmentation model has no guarantee under bad imaging conditions or adverse weather such as rain and fogs.     

From the perspective of improve robustness against artificially generated adversarial perturbation, a series of attack and defense methods have been developed. As a special case of the classification tasks, the results of segmentation model will also be degraded by the classical adversarial attacks for image classification tasks, such as FGSM~\cite{goodfellow2014explaining}, PGD~\cite{madry2017towards} and their variations~\cite{wong2020fast,zhang2022revisiting,croce2020reliable}. Besides, another line of work~\cite{gu2022segpgd,agnihotri2023cospgd} has explored the difference between semantic segmentation and image classification to design task-specific segmentation attack methods and generate more effective adversarial examples. As one of most effective defense strategy, Adversarial Training (AT)~\cite{kurakin2016adversarial,tramer2017ensemble,song2018improving} addresses the vulnerability of segmentation model by incorporating the adversarial example during the training process, which can be further formulated as the minimax optimization problem. In addition to few AT based methods~\cite{xu2021dynamic} for semantic segmentation, several works also apply teacher-student structure~\cite{bar2019robustness} and multitask learning~\cite{mao2020multitask} to improve the robustness of segmentation model.

To improve the limited performance of semantic segmentation under extreme weather, recently proposed methods have explored various techniques for low-level tasks. Take the rainy weather as an example, Single Image Deraining (SID)~\cite{fu2019lightweight,deng2020detail} aims to remove the degradation noise from the input rainy images and retains as much context details as possible, which could be embedded as the low-level preprocessing procedure to benefit the downstream segmentation tasks. In comparison with the optimization based methods~\cite{kang2011automatic, zhang2018density, li2018recurrent, xu2021intensity}, varieties of deep learning based methods~\cite{fu2017removing,hu2019depth,li2019heavy} explore different network structures to obtain significant performance based on massive training data. Besides, several methods~\cite{li2018recurrent,zhang2020beyond} have also incorporated the high-level semantic knowledge as efficient feedback to facilitate the deraining process.

Whereas, the above methods essentially focus on eliminating a specific influence factor such as bad weather or adversarial attacks to enhance the adaptability or robustness of segmentation model in real-world applications, while implying no uniform understanding of these degradation factors which influence the performance of high-level tasks. To be general, the environmental degradation phenomenon and artificially introduced adversarial perturbation share similar principles for semantic segmentation tasks, and could be regarded as some specific form of degradation factors added to the input of the segmentation input. From this new perspective, we make our attempt to design a novel framework to jointly handle both types of degradation factors without introducing additional network parameters or task-specific loss functions.

Firstly, we introduce the Naive Adversarial Training (NAT) framework, which improves the robustness of segmentation model based on AT while handling the rain streaks by embedding extra image deraining module. Whereas, separately removing the rain steaks and defending adversarial perturbation will deteriorate the derain model and introduce residual perturbation to the output of the derain model, which finally affects the downstream tasks. Inspired by the idea which designs specialized transformation module concatenated to the original classification model to defend adversarial examples, we here propose to transfer the robustness of segmentation model to the derain model, and design the Preprocessing Enhanced Adversarial Robust Learning (PEARL) framework to simultaneously deal with both adversarial perturbation and rain streaks. Moreover, as opposed to the Negative Adversarial Attack (NAA), we propose the Auxiliary Mirror Attack (AMA) to introduce "positive" information prior of the adversarial attack to the supervised training of derain model, which enhances the defense capability of derain model and improves the segmentation results eventually. Experimentally, we conducted extensive experiments and ablation studies to demonstrate the performance improvement of both derain and segmentation results with quantitative and visualization results. Moreover, we also verify the generalization performance of our framework across different datasets.  

The main contributions of this paper are summarized as follows.
\begin{itemize}
	\item To the best of our knowledge, we make the first attempt delving into downstream semantic segmentation tasks influenced by both natural degradation factor (e.g., rain streaks) and artificially generated degradation factors (e.g., adversarial attacks), and significantly improve the downstream task performance under bad weather while retaining the robustness against adversarial attacks.  
	\item In contrast with the proposed Naive Adversarial Training (NAT) framework, we introduce our Preprocessing Enhanced Adversarial Robust Learning (PEARL) framework to transfer the robustness of segmentation model, aiming to obtain high-performance robust derain model, which can effectively eliminate the influence of both degradation factors on the segmentation attacks.    
	\item We design another Auxiliary Mirror Attack (AMA) as opposed to the Negative Adversarial Attack (NAA) to embed positive perturbation to the proposed PEARL framework, which facilitates the robust learning to improve the defense capability and leads to better segmentation results. 
	\item Extensive experimental results and ablation studies on different derain and segmentation models have demonstrated the effectiveness of PEARL framework and AMA module to enhance the robustness against various degradation attacks and improve the deraining performance. Besides, we also verify the generalization performance of our proposed framework based on different datasets.
\end{itemize}

\section{Related Works}

\subsection{Image Deraining for Semantic Segmentation}

\textbf{High-level Segmentation Task.} As a specific form of pixel-level dense classification tasks, semantic segmentation have been well developed to explore the contextual dependencies and capture the long-range relationship. Chen et.al.~\cite{chen2017deeplab} proposed DeepLab, which introduces the atrous convolution for explicit resolution control and uses spatial pyramid pooling for multi-scale objective segmentation. Then they developed the DeepLabv3~\cite{chen2017rethinking} for further improvements to atrous convolution and atrous spatial pyramid pooling modules and also incorporates image-level features for global context. Zhao et. al. ~\cite{zhao2017pyramid} presents PSPNet which exploits global context information and pyramid pooling module to improve segmentation performance on various scene parsing datasets. Xie et.al.~\cite{xie2021segformer} proposed SegFormer to unifiy Transformers with lightweight MLP decoders and achieves state-of-the-art performance with simple and efficient design. Practically speaking, the above methods have spared efforts to work on degradation-free scenes, which may faces serious performance decrease under adverse weather. 

\textbf{Low-level Deraining Task.} Single Image Deraining (SID)~\cite{fu2019lightweight,deng2020detail} has been well developed to deal with different rain streaks and improve the downstream tasks for practical applications.  Typically, Li et.al.~\cite{li2018recurrent} proposed RESCAN to incorporate dilated convolutional neural networks and recurrent neural networks to remove rain streaks in multiple stages. Ren et.al.~\cite{ren2019progressive} constructs a better and simpler baseline deraining network, called PReNet, which provides consistent improvements of the architecture and loss functions. Zamir et.al.~\cite{zamir2021multi} introduces a multi-stage architecture called MPRNet to progressively learn restoration functions for degraded inputs and balances the competing goals of spatial details and high-level contextualized information in image restoration tasks. Recently, Valanarasu et. al.~\cite{valanarasu2022transweather} proposed transformer-based model with a single encoder and a decoder that can restore an image degraded by any weather condition. Besides, a line of works~\cite{li2019heavy, jiang2020multi} also explore the high-level semantic information, such as the detection and segmentation results, to guide the optimization of deraining process. Note that our propose training framework of image deraining implies no explicit requirements of the network structure or design of loss functions, which makes it capable to incorporate recent-proposed methods to obtain higher performance.

\subsection{Adversarial Attacks and Defenses}

\textbf{Adversarial Attacks.} It has been investigated~\cite{arnab2018robustness,xie2017adversarial} that the segmentation model also shows vulnerability to these artificially introduced adversarial examples. Generally speaking, the adversarial attacks include two categories, i.e., black-box attacks~\cite{papernot2017practical} and white-box~\cite{athalye2018robustness} attacks. Here we focus on the gradient-based white-box attack which is capable to access full knowledge of the model under attack (known as target model), and generated imperceptible perturbations by computing the gradient of target model. Several commonly used adversarial attack methods include FGSM~\cite{goodfellow2014explaining} and PGD~\cite{madry2017towards}, which generated single-step and multi-step perturbation for the input image. Based on two basic attacks, different attack methods have been explored by introducing practical techniques~\cite{dong2018boosting,zhang2022revisiting}. kurakin et.al.~\cite{kurakin2018adversarial} proposed BIM attack and demonstrates that machine learning systems are vulnerable to adversarial examples even in physical world scenarios. Carlini et.al.~\cite{carlini2017towards} challenges the effectiveness of defensive distillation and introduces the optimization based attack method denoted as CW. In addition to the above general-purpose attacks, several works~\cite{gu2022segpgd,agnihotri2023cospgd} also conduct impressive investigation on the robustness of segmentation and introduce effective improvements of the PGD attack, which also shows its necessity of training robust segmentation model for better defense against the adversarial degradation factors.

\textbf{Adversarial Defense.} Generally speaking, Adversarial Training (AT)~\cite{kurakin2016adversarial,tramer2017ensemble} trains the model to defend the adversarial example by minimizing the attack objective, which also make it more time-consuming due to generation of adversarial example and tasks more epochs to converge. Whereas, few works have explored the effectiveness of AT on the segmentation model tasks. Practically, by setting additional branches in the target model during training and dividing pixels into multiple branches, Xu et.al.~\cite{xu2021dynamic} proposed DDC-AT for improving the robustness of deep neural networks on semantic segmentation tasks. In addition, another branch of defense methods have investigated different transformations, such as image compression and pixel deflection~\cite{prakash2018deflecting, song2017pixeldefend}, embedded to preprocess the input, thus remove the adversarial perturbation. There also has been a lack of research in recent years that have continued to investigate this direction. Instead of directly using AT to handle different degradation factors, we here employ the preprocessing based idea to construct the robust learning process with embedded derain model, which is supposed to jointly handle both rain streaks and adversarial attacks. Besides, our framework retains more flexibility to be further improved with task-specific model design and additional loss functions.  

\begin{figure*}[h!]
	\begin{center}
		\renewcommand\arraystretch{1.0}
		\begin{tabular}{@{\extracolsep{0.2em}}c@{\extracolsep{0.2em}}c@{\extracolsep{0.2em}}c@{\extracolsep{0.2em}}}
			&\includegraphics[height=7.58cm,width=17cm,trim=0 0 0 0,clip]{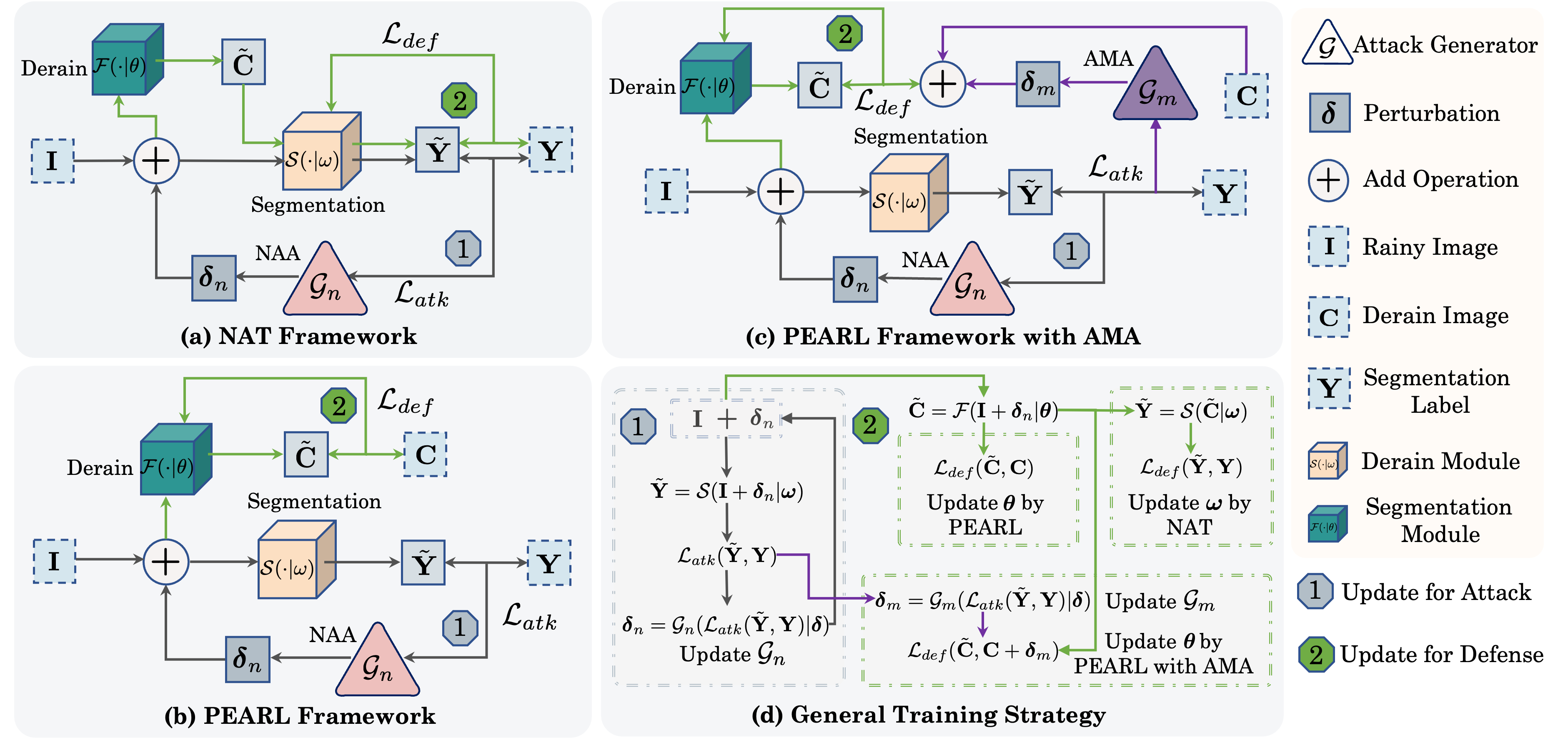}&\\
			\end{tabular}
		\end{center}
		\vspace{-0.2cm}
		\caption{ The first three subfigures illustrate the Naive Adversarial Training (NAT) training framework for handling rain streaks and adversarial attacks for image segmentation model, our Preprocessing Enhanced Adversarial Robust Learning (PEARL) framework and its whole pipeline with proposed Auxiliary Mirror Attack (AMA) technique. The last subfigure describes the training strategy for NAT, PEARL and PEARL with AMA. We use gray, green and purple lines to denote the optimization cycle of attack, defense and additional flow introduced by the AMA module. }\label{pipeline}
	\end{figure*}

\section{Proposed Method}

In this section, we first provide simplified problem definition of different degradation attacks factors and AT to derive the Naive Adversarial Training (NAT) framework for improving robustness of image segmentation model. With analysis of the limitation of NAT, we further propose our Preprocessing Enhanced Adversarial Robust Learning (PEARL) framework with designed Auxiliary Mirror Attack (AMA) generator, by which simultaneously remove the rain streaks and improve the robustness to defend downstream adversarial attacks.  

\subsection{Naive Adversarial Training Framework Against Degradation Attacks}

In this work, we consider an image segmentation model $\mathcal{S}(\cdot|\boldsymbol{\omega})$ parameterized by $\boldsymbol{\omega}$. Given a training dataset $\mathcal{D}_{\textrm{tr}}$ with labeled data pairs, the segmentation output can be represented as $\tilde{\mathbf{Y}} =\mathcal{S}(\mathbf{C}|\boldsymbol{\omega})$, where $\mathbf{C}$ denotes the input image, and $\tilde{\mathbf{Y}}$ denotes the output label of segmentation. Therefore, this downstream task aims to optimize the following objective: $\min_{\boldsymbol{\omega}}\mathcal{L}_{\textrm{seg}}(\tilde{\mathbf{Y}},\mathbf{Y})$, where $\mathbf{Y}$ denotes the groundtruth label of segmentation.

Typically speaking, the adversarial attack is supposed to deteriorate the output label of segmentation model by introducing visually imperceptible perturbation, i.e., $\boldsymbol{\delta}$ to the input image, which can be reformulated as  

\begin{equation}
	\centering
	\boldsymbol{\delta}=\underset{\boldsymbol{\delta},\|\boldsymbol{\delta}\|_p \leq \epsilon}{\arg \max }\mathcal{L}_{atk}(\mathcal{S}(\mathbf{C}+\boldsymbol{\delta}|\boldsymbol{\omega}),\mathbf{Y} ), \label{delta_form}
\end{equation}
where $\boldsymbol{\delta}$ is usually bounded with $\epsilon$-toleration $\ell_{p}$-norm, $\boldsymbol{\delta} \in [0,1]$, and $\mathcal{L}_{atk}$ is the adversarial loss to measure the distance between generated degraded example and ground truth. Typically, we could consider the same form of $\mathcal{L}_{\textrm{seg}}$ to define the adversarial loss function. Based on the above formulation, when we apply $K$-step PGD method to generate the adversarial example, the perturbation at $k$-th step can be denoted as 
\begin{equation}
	\centering
	\boldsymbol{\delta}^{k+1} \leftarrow \Pi_{\epsilon}(\boldsymbol{\delta}^k + \alpha \cdot sgn (\nabla_{\boldsymbol{\delta}} \mathcal{L}_{atk}(\mathcal{S}(\mathbf{C}+\boldsymbol{\delta}^{k}|\boldsymbol{\omega})),\mathbf{Y} )),\label{delta_update}
\end{equation}
where $k=0,\dots,K-1$, $\alpha$ is the step size for perturbation generation, $\Pi_{\epsilon}(\cdot)$ and $sgn(\cdot)$ denotes the projection operation and element-wise $sign$ operation, respectively. The initial perturbation $\boldsymbol{\delta}^{0}$ is sampled from uniform distribution $U(-\epsilon, \epsilon)$. In the following, we use $\boldsymbol{\delta}_{n}$ to represent the adversarial attack $\boldsymbol{\delta}^{K}$ generated by a specific Negative Adversarial Attack (NAA) generatior  denoted as  $\boldsymbol{\delta}_{n}=\mathcal{G}_{n}(\mathcal{L}_{atk}(\mathcal{S}(\mathbf{C}+\boldsymbol{\delta}|\boldsymbol{\omega}),\mathbf{Y} )|\boldsymbol{\delta})$, (e.g., PGD), in order to distinguish them from the auxiliary mirror attacks we introduced later.

As it is mentioned above, AT have been extensively investigated to defend the adversarial attacks by solving the following minimax optimization problem  

\begin{equation}
	\centering
	\underset{\boldsymbol{\omega}}{\operatorname{minmize} } \ \mathbb{E}_{(\mathbf{C},\mathbf{Y}) \in \mathcal{D}_{\textrm{tr}}}  \Big [ \underset{\boldsymbol{\delta},\|\boldsymbol{\delta}\|_p \leq \epsilon}{\operatorname{maximize}}\  \mathcal{L}_{atk}(\mathcal{S}(\mathbf{C}+\boldsymbol{\delta}_{n}|\boldsymbol{\omega}),\mathbf{Y} ) \Big]. \label{naive_at}
\end{equation}
By alternatively optimizing $\boldsymbol{\omega}$ and generating new perturbation $\boldsymbol{\delta}_{n}$ with $\mathcal{G}_{n}(\cdot|\boldsymbol{\delta})$, the robustness of segmentation against adversarial samples generated by different types of NAAs can be consistently improved. The objective of adversarial defense for the segmentation model is denoted as $\mathcal{L}_{def}$, which is usually defined as the same form of $\mathcal{L}_{atk}$.  

Whereas, we consider more general setting where the manually designed adversarial attack is essentially regarded as one of the specific form of degradation attacks factors. In this case, we are allowed to consider various degradation factors such as inevitable noises caused by extreme weather, e.g., rain and fog, which are prior existing parts of the original input $\mathbf{C}$. Here we consider the degraded factors as rainstreaks, and denote the degraded rainy image as $\mathbf{I}$ when the rain streaks exist in the input.

\begin{figure}[htb]
	\centering \begin{tabular}{@{\extracolsep{-0.1em}}c@{\extracolsep{0.3em}}c@{\extracolsep{0.3em}}c@{\extracolsep{0.3em}}c@{\extracolsep{0.15em}}}
		\includegraphics[height=0.065\textheight,width=0.135\textwidth]{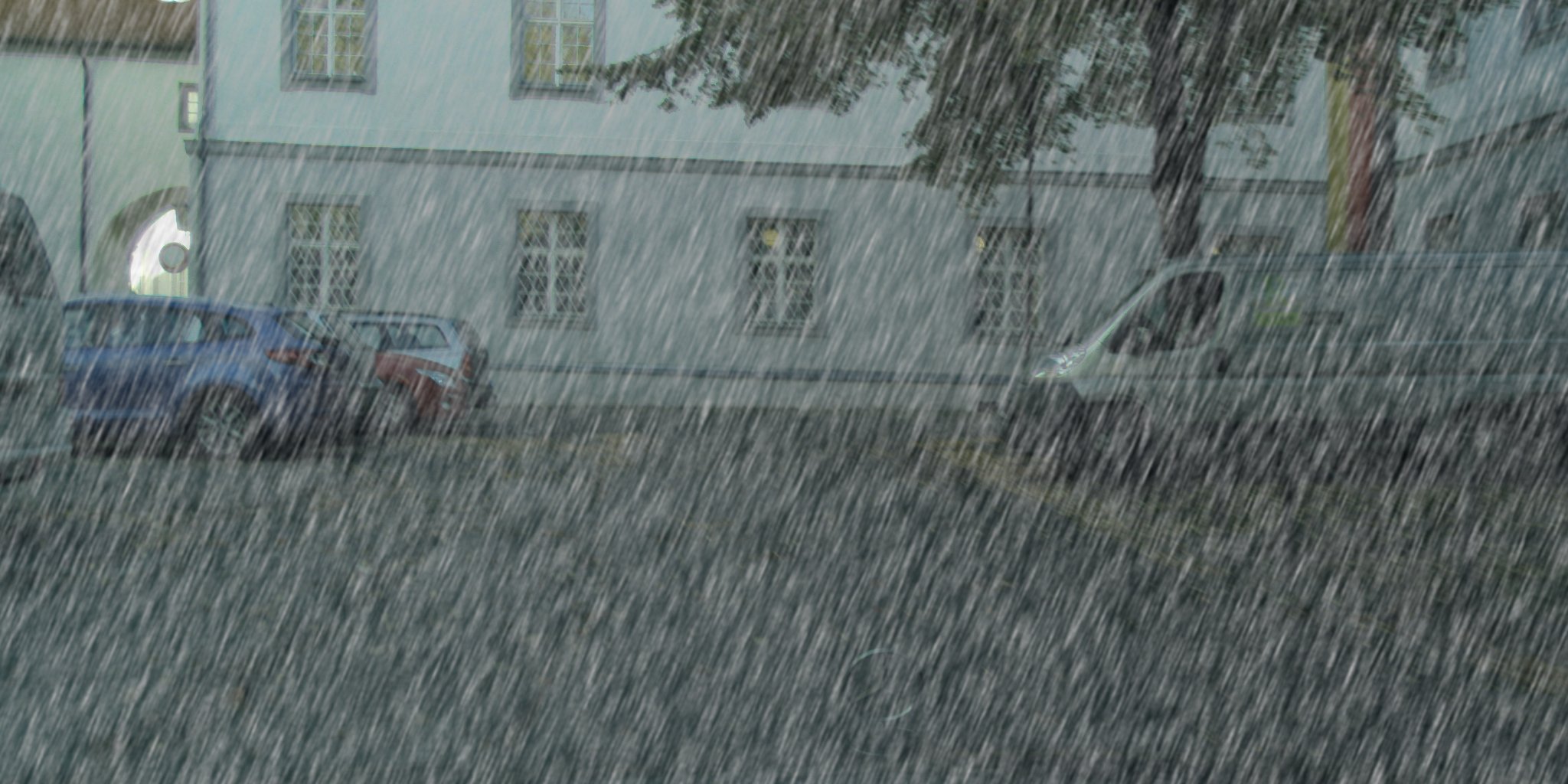}
		&\includegraphics[height=0.065\textheight,width=0.135\textwidth]{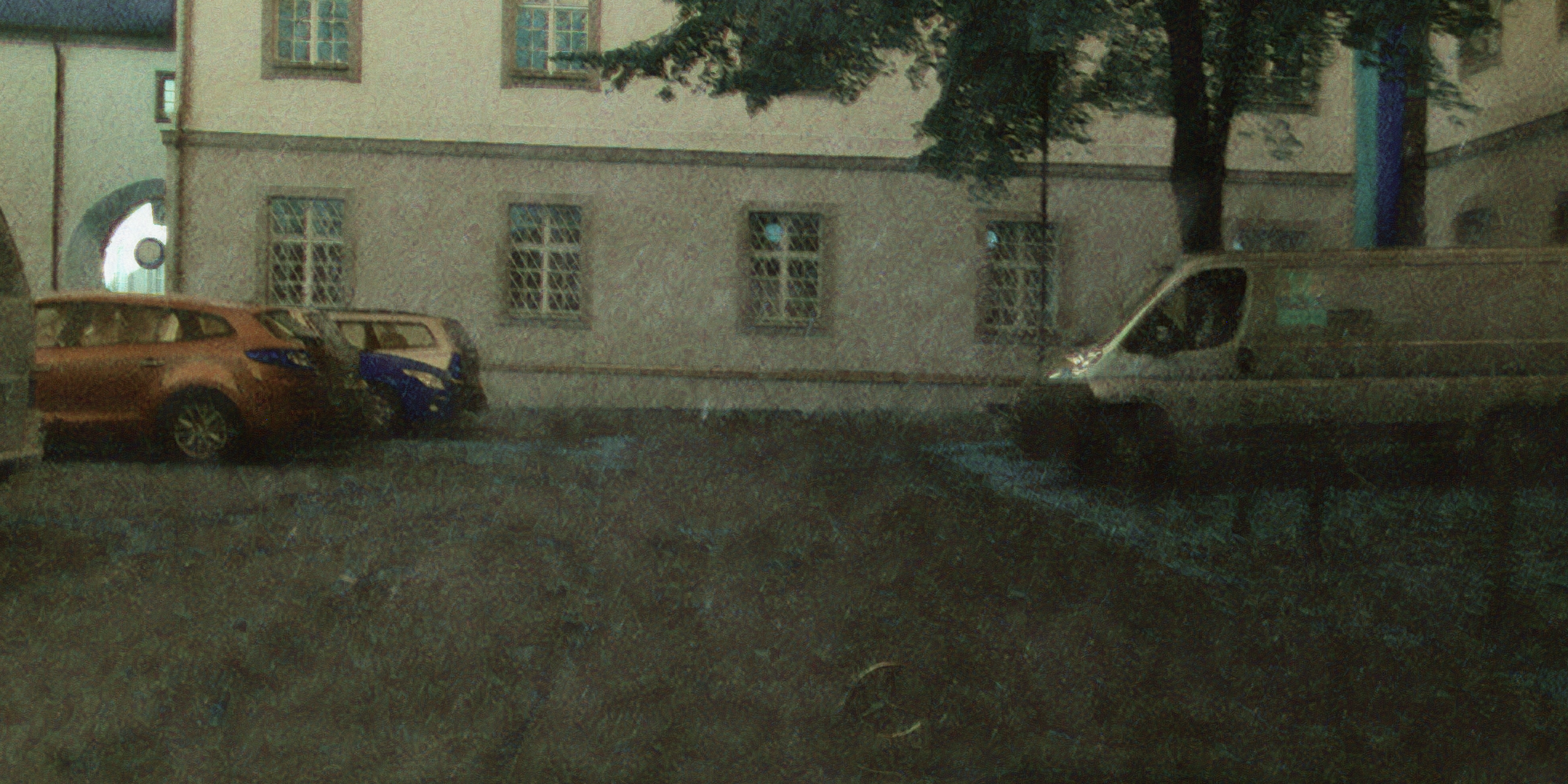}
		& \includegraphics[height=0.065\textheight,width=0.135\textwidth]{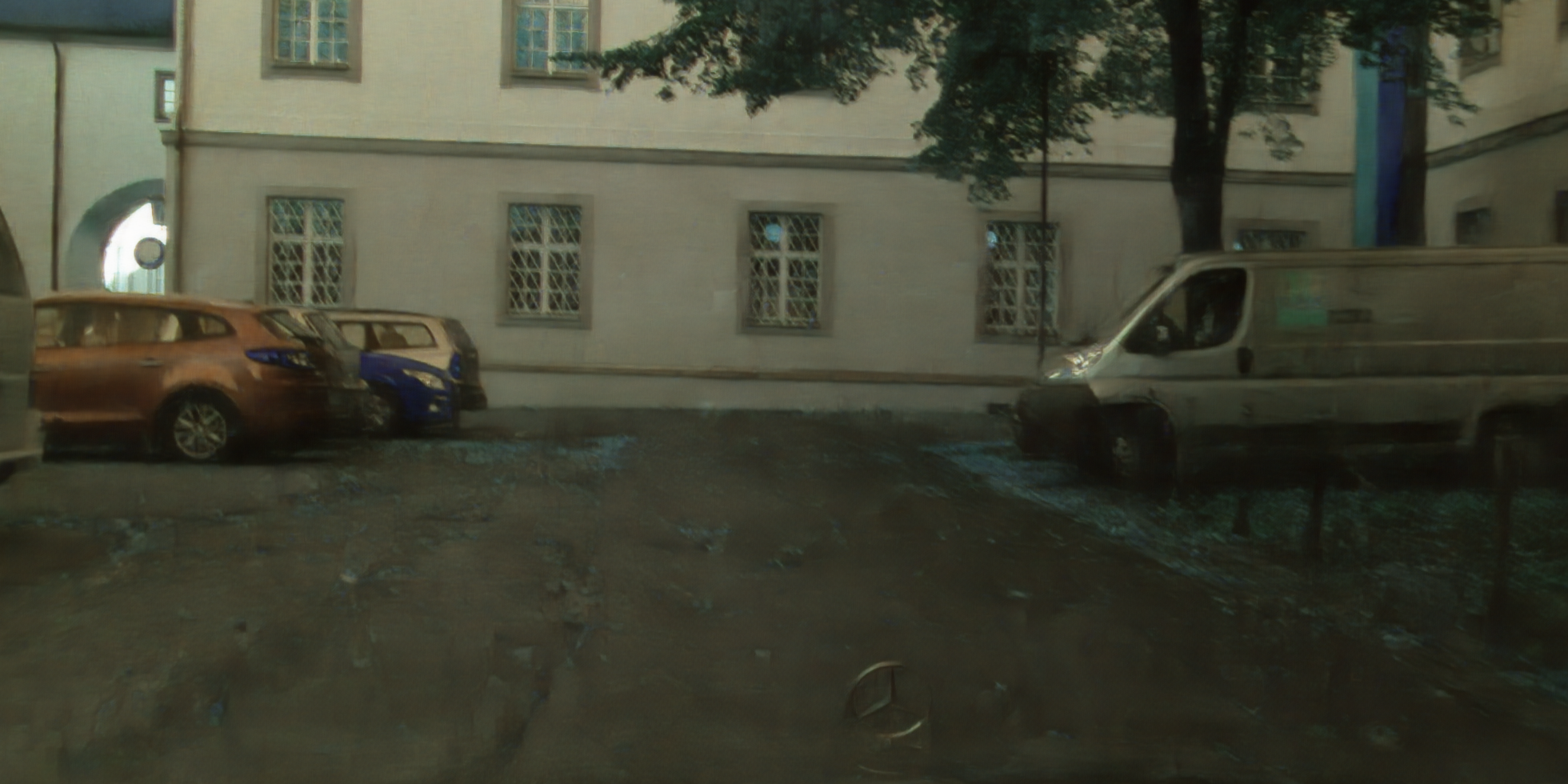}\\
		\specialrule{0em}{-1pt}{-1pt}
			\footnotesize (a) $\mathbf{I} + \boldsymbol{\delta}_{n}$ & \footnotesize (c) $\mathcal{F}(\mathbf{I} + \boldsymbol{\delta}_{n}|\boldsymbol{\theta})$  &  \footnotesize (e) $\mathcal{F}(\mathbf{I} + \boldsymbol{\delta}_{n}|\boldsymbol{\theta})$ \\
		\includegraphics[height=0.065\textheight,width=0.135\textwidth]{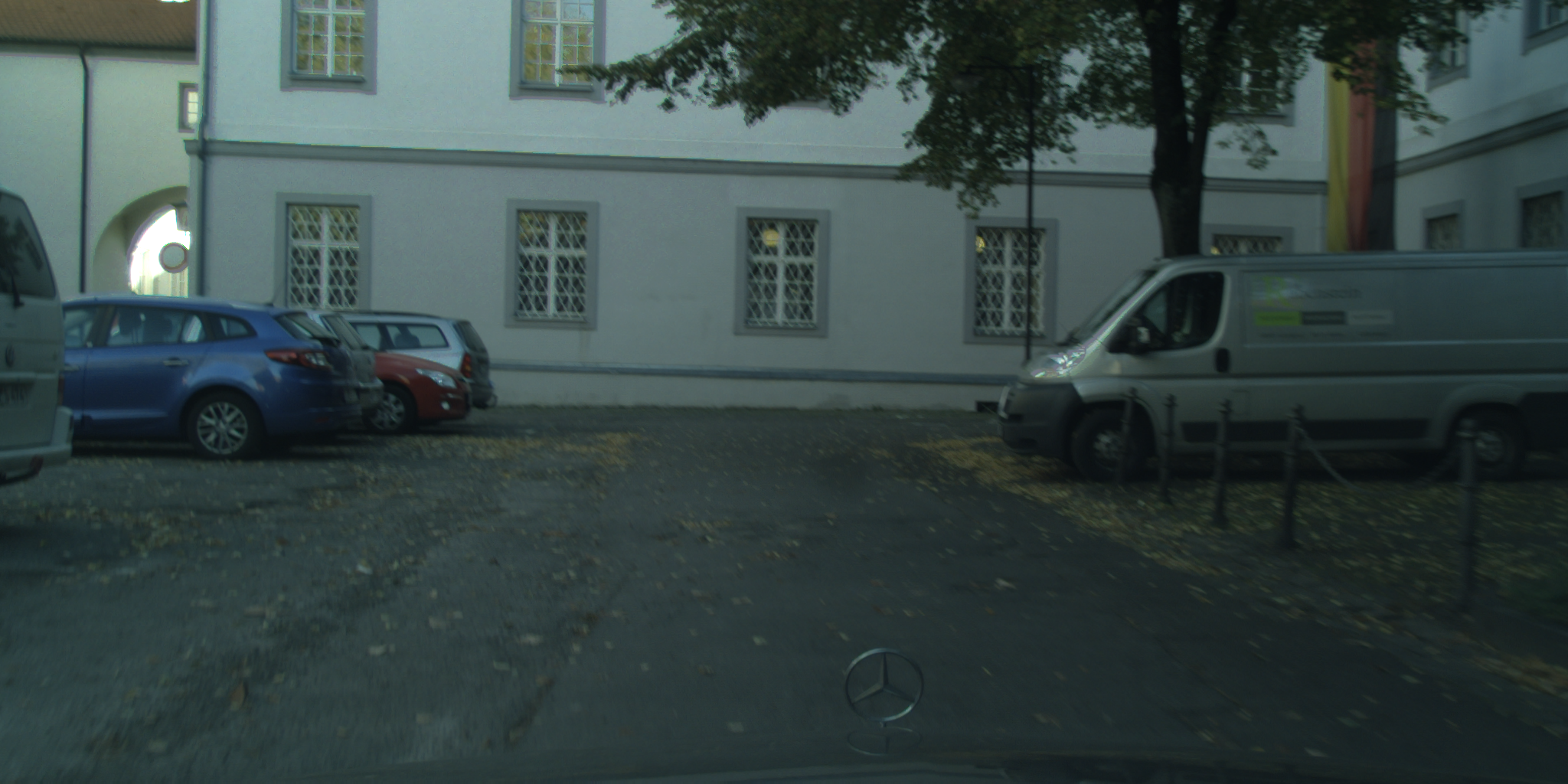}
		&\includegraphics[height=0.065\textheight,width=0.135\textwidth]{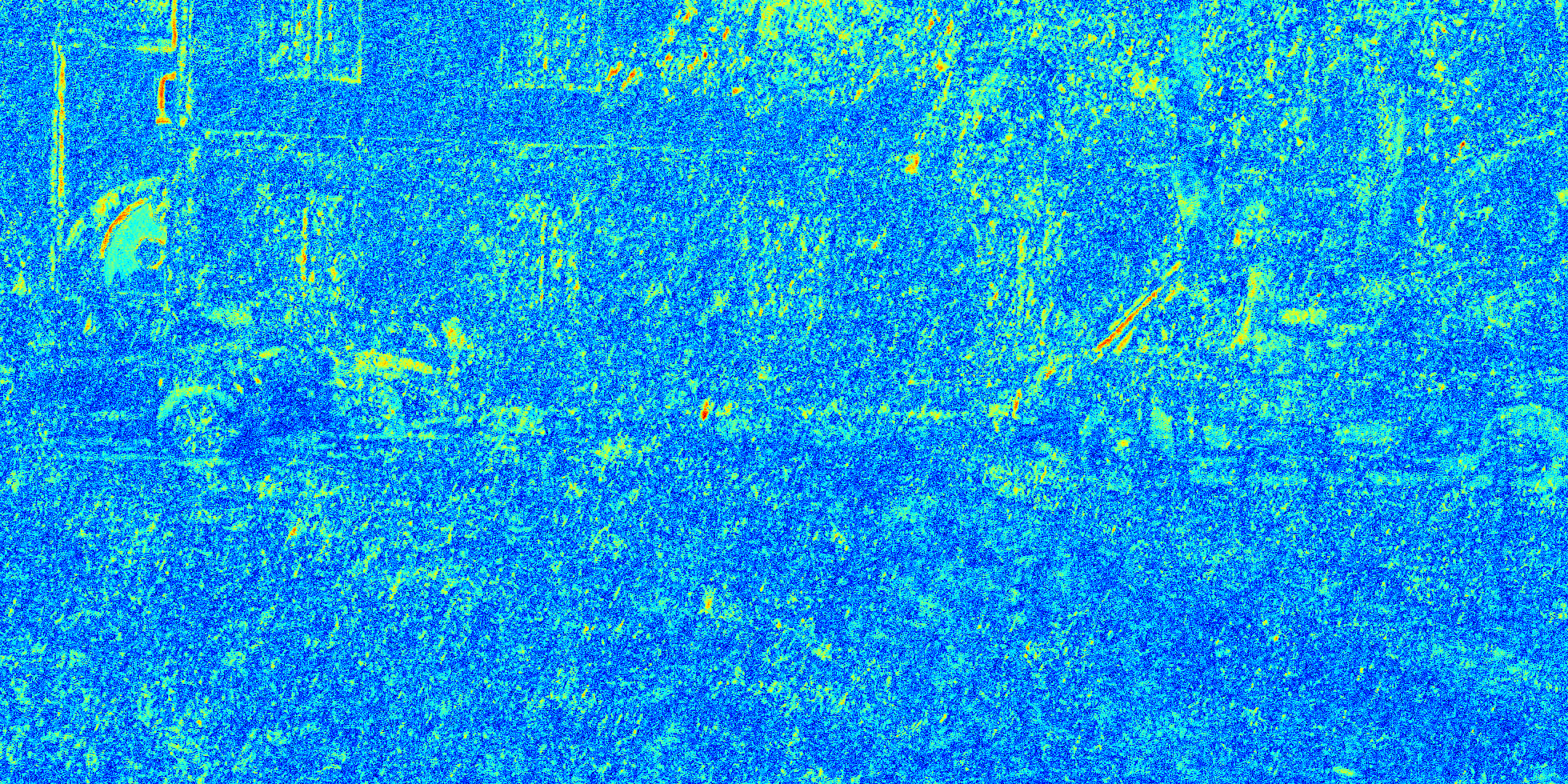}
		& \includegraphics[height=0.065\textheight,width=0.135\textwidth]{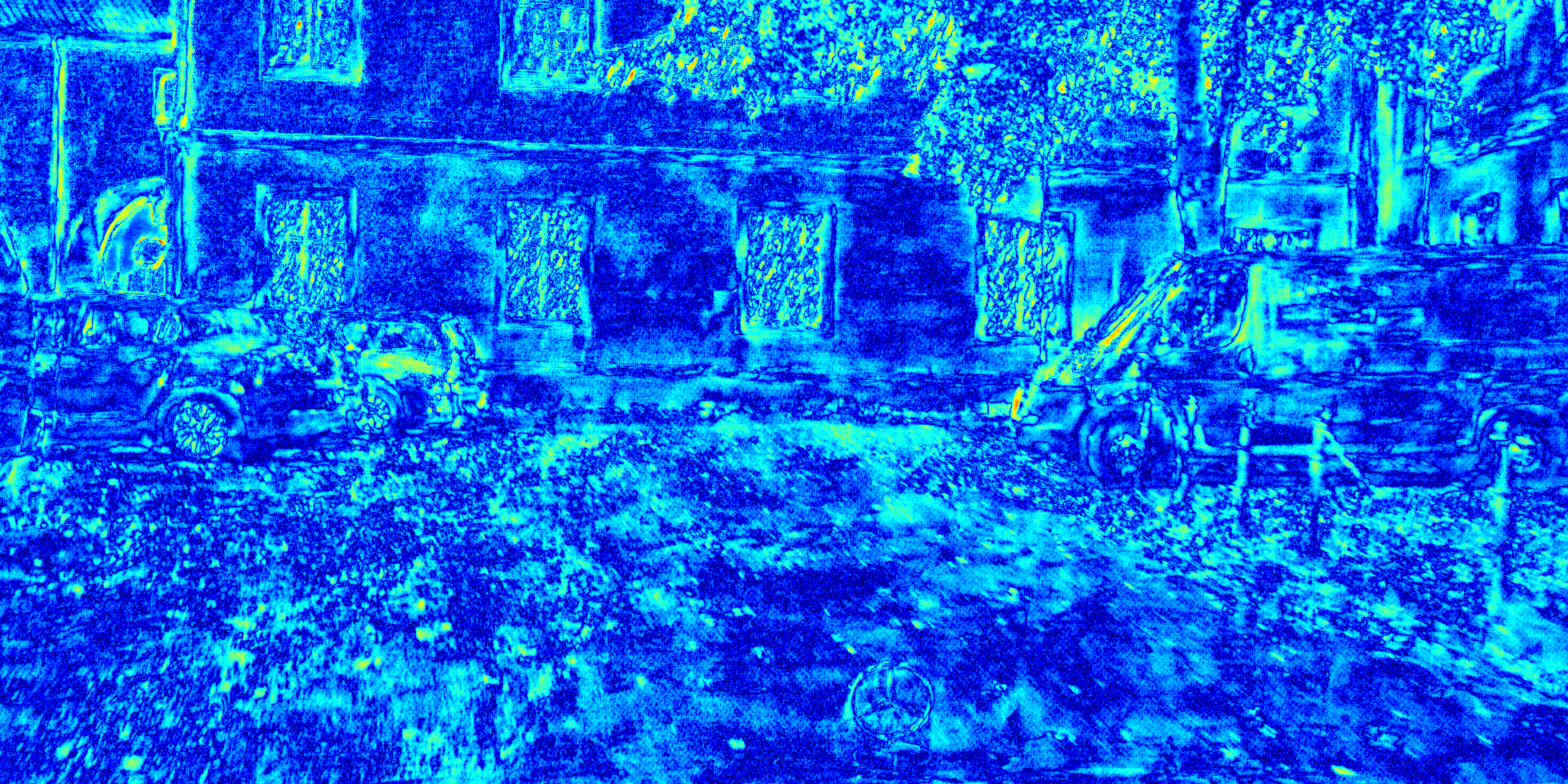} \\
				\specialrule{0em}{-1pt}{-1pt}
			\footnotesize (b) Clean GT & \footnotesize  (d) Pretrained $\mathcal{F}(\cdot|\boldsymbol{\theta})$ & \footnotesize (f) $\mathcal{F}(\cdot|\boldsymbol{\theta})$ of PEARL \\
		
		%
	\end{tabular}
	\vspace{-0.2cm}
	\caption{We compare the processed heat maps of pretrained derain model and our proposed framework to show the difference between derain results and groundtruth with both rain streaks and BIM attack (${\epsilon} = 4/255$).}
	\label{fig:heat_map}
\end{figure}

To alleviate the negative impact of both degradation attacks, i.e., the rain streaks and adversarial perturbation, we first propose the Naive AT (NAT) framework, which can be illustrated in Fig.~\ref{pipeline}(a). It first embed pretrained derain model (denoted as $\mathcal{F}(\cdot|\boldsymbol{\theta})$, where $\boldsymbol{\theta}$ denotes the parameters of derain model) to remove the rain streaks, and retain the robust segmentation model to handle the adversarial attacks for downstream tasks. Whereas, since the derain model encompasses little prior of the adversarial distribution, the perturbations added in the rainy image may deteriorate the deraining results seriously. In Fig.~\ref{fig:heat_map}, we analyze the heatmap of deraining results processed with pretrained derain model and the one trained with our proposed framework.  As it can be observed, when the perturbation generated to attack the segmentation model exists in the rainy image $\mathbf{I}$, it will also severely degrade the derain result and leave imperceptible disturbance in the output, which is a mix of multiple degradation factors. Consequently, the residual perturbation and rain streaks left in $\tilde{\mathbf{C}}=\mathcal{F}(\mathbf{I} + \boldsymbol{\delta}_{n}|\boldsymbol{\theta})$ also results in the performance decrease of robust segmentation model to defend the adversarial attacks, which increases the difficulty of AT based framework. As one of the most significant contributions, we fully explore the potential capability of embedded derain model, and design the following Preprocessing Enhanced Adversarial Robust Learning (PEARL) framework to effectively defend both manually designed attack and natural degradation attacks. 
       
\subsection{Preprocessing Enhanced Adversarial Robust Learning (PEARL) Framework}

To be general, the decomposition mapping function of derain model could be rationally reformulated as: $\mathbf{I} = \mathbf{C} + \mathbf{R}$, where $\mathbf{C}$ and $\mathbf{R}$ represent the clean background and rain layers extracted from the degraded input. According to the above formulation of adversarial attacks, the degraded input with the adversarial perturbation is denoted as $\mathbf{I} + \boldsymbol{\delta}_{n}$. In this case, the introduced adversarial perturbation may be regarded as certain noises added to the clean image to some extent.  Typically, the denoising task aims to learn the following denoising mapping function: $\tilde{\mathbf{C}} \to \mathbf{C}$, where $\tilde{\mathbf{C}}$ and $\mathbf{C}$ denote the input image with noises and clean image. From this perspective, the rain streaks and adversarial perturbation can be all treated as noises, which could be further removed by embedding  denoisers.

 Inspired by the preprocessing based methods~\cite{gu2014towards, liao2018defense} which remove the adversarial noise by designing specific transformation modules, we here make our attempt to transfer the robustness of segmentation model against adversarial attacks to the embedded derain model.   In this case, we no longer follow the AT based formulation in Eq.~\eqref{naive_at} to implement NAT framework in Fig.~\ref{pipeline}(a), and introduce the following Preprocessing Enhanced Adversarial Robust Learning (PEARL) framework:
	
\begin{equation}
	   	\begin{aligned}
	&\underset{\boldsymbol{\theta}}{\operatorname{min} } \   \mathcal{L}_{def}(\mathcal{F}(\mathbf{I} + \boldsymbol{\delta}_{n}|\boldsymbol{\theta}),\mathbf{C} ) \\
	&s.t. \boldsymbol{\delta}\in \underset{\boldsymbol{\delta},\|\boldsymbol{\delta}\|_p \leq \epsilon}{\operatorname{argmax}}\  \mathcal{L}_{atk}(\mathcal{S}(\mathbf{I}+\boldsymbol{\delta}_{n}|\boldsymbol{\omega}),\mathbf{Y} ) ,
		\label{pirt}
	\end{aligned}
\end{equation}
where $\mathcal{L}_{def}$ could be specified as the objective function of derain model. Besides, we can introduce additional regularization terms to $\mathcal{L}_{def}$ as task priors of downstream segmentation tasks based on the output of derain model.  As described in Fig.~\ref{pipeline}(b) and Eq.~\eqref{pirt}, the degraded example is still generated by adding adversarial perturbation to the rainy image, while we replace the outer minimization optimization of $\mathcal{S}(\cdot|\boldsymbol{\omega})$ in Eq.~\eqref{naive_at} with training derain model to learn the following decomposition mapping function: 
\begin{equation}
	\centering
	\mathbf{I}+\boldsymbol{\delta}_{n}\to \tilde{\mathbf{C}} +( \mathbf{R} +\boldsymbol{\delta}_{n}),
	\label{dempose_map}
\end{equation}
where $\tilde{\mathbf{C}}$ is approximated by minimizing $\mathcal{L}_{def}(\tilde{\mathbf{C}},\mathbf{C} )$.

Practically, we simply restore the segmentation weights pretrianed based on the clean images, and optimize the negative attack generator and derain model in an alternative manner. The derain model trained with PEARL framework is supposed to jointly remove the rain streaks and adversarial noise, thus make the derain results, i.e., $\mathcal{F}(\mathbf{I} + \boldsymbol{\delta}_{n}|\boldsymbol{\theta})$, closer to the clean image. Consequently, the preprocessed results have weakened the negative influence of both degradation attack factors, which also enhance the downstream segmentation tasks to a great extent. In the next subsection, to make the utmost of generated adversarial noise based on the inner maximization, we introduce another  auxiliary mirror attack to mimic the deterioration process of adversarial attack, and incorporate the generated positive perturbation to facilitate the noise decomposition of derain model.

\begin{table*}[h!]
	\renewcommand{\arraystretch}{1.2}
	\setlength{\tabcolsep}{1 mm}
	\caption{Evaluation results with both natural and artificial degradation factors on synthesized Cityscapes dataset. Adversarial attack is generated by BIM ($K=3,5,10$), PGD10 and CW, respectively. We report the defense results with perturbation value $\epsilon=8/255$, and more results for the perturbation $\epsilon=4/255$ can be found in the supplementary materials.}	\label{table_all_attack}
	\vspace{-0.2cm}
	\centering
	\begin{tabular}{c|ccccccccccccccc}
		\toprule[1.3pt]
		\toprule[1.1pt]
		\multirow{2}{*}{Methods} & \multicolumn{3}{c|}{Rain+BIM3}                   & \multicolumn{3}{c|}{Rain+BIM5}                   & \multicolumn{3}{c|}{Rain+BIM10}                  & \multicolumn{3}{c|}{Rain+PGD10}                  & \multicolumn{3}{c}{Rain+CW}                      \\
		& mIoU           & allAcc         & \multicolumn{1}{c|}{PSNR}           & mIoU           & allAcc         & \multicolumn{1}{c|}{PSNR}           & mIoU           & allAcc         & \multicolumn{1}{c|}{PSNR}           & mIoU           & allAcc         & \multicolumn{1}{c|}{PSNR}           & mIoU           & allAcc         & PSNR           \\ \midrule
		Seg         & 2.81           & 31.36          & \multicolumn{1}{c|}{17.44}          & 2.46           & 27.61          & \multicolumn{1}{c|}{17.39}          & 2.41           & 27.40          & \multicolumn{1}{c|}{17.39}          & 2.42           & 27.87          & \multicolumn{1}{c|}{17.39}          & 1.80           & 21.74          & 17.41          \\
		Robust Seg                        & 2.16           & 38.32          & \multicolumn{1}{c|}{17.37}          & 2.08           & 38.02          & \multicolumn{1}{c|}{17.28}          & 2.06           & 37.93          & \multicolumn{1}{c|}{17.23}          & 2.05           & 37.91          &  \multicolumn{1}{c|}{17.22}          & 2.09         & 38.04          & 17.29       \\
		Derain + Seg                        & 9.31           & 38.28          & \multicolumn{1}{c|}{29.46}          & 3.79           & 20.02          & \multicolumn{1}{c|}{28.83}          & 1.90           & 13.56          & \multicolumn{1}{c|}{28.24}          & 1.92           & 12.84          & \multicolumn{1}{c|}{28.25}          & 3.12           & 15.13          & 28.83          \\ \hline
		NAT                               & 38.39          & 85.03          & \multicolumn{1}{c|}{29.78}          & 34.31          & 82.00          & \multicolumn{1}{c|}{28.80}          & 31.37          & 79.24          & \multicolumn{1}{c|}{28.08}          & 31.31          & 79.10          & \multicolumn{1}{c|}{28.08}          & 34.57          & 82.12          & 28.68          \\
		PEARL(Ours)                       & 47.81          & 88.80          & \multicolumn{1}{c|}{\textbf{32.62}} & 44.70          & 87.10          & \multicolumn{1}{c|}{\textbf{32.31}} & 41.03          & 83.86          & \multicolumn{1}{c|}{\textbf{31.86}} & 41.69          & 84.44          & \multicolumn{1}{c|}{\textbf{31.88}} & 46.16          & 87.12          & \textbf{32.30} \\
		+AMA(Ours)                  & \textbf{48.55} & \textbf{88.81} & \multicolumn{1}{c|}{32.56}          & \textbf{46.14} & \textbf{87.55} & \multicolumn{1}{c|}{32.21}          & \textbf{43.75} & \textbf{85.95} & \multicolumn{1}{c|}{31.74}          & \textbf{44.60} & \textbf{86.34} & \multicolumn{1}{c|}{31.77}          & \textbf{47.73} & \textbf{87.84} & 32.20        \\ 
		
		\bottomrule[1.1pt]
		\bottomrule[1.3pt]
		
	\end{tabular}
	\vspace{-0.1cm}
\end{table*}

\subsection{Auxiliary Mirror Attack (AMA)}

In this subsection, we propose another enhancement technique to assist the optimization of derain model and further improve the performance of downstream segmentation tasks. Based on previous definition of $\mathcal{G}_{n}$, the generated perturbation $\boldsymbol{\delta}_{n}=\mathcal{G}_{n}(\mathcal{L}_{atk}(\tilde{\mathbf{Y}},\mathbf{Y})|\boldsymbol{\delta})$ is added to the input of derain model to involve the degraded attack, as illustrated in Fig.~\ref{pipeline}(b). By minimizing the outer objective $\mathcal{L}_{def}$ in Eq.~\eqref{pirt}, we have injected the noise distribution of adversarial attack into the derain model such that $\mathcal{F}(\cdot|\boldsymbol{\theta})$ generalize to this decomposition mapping task and minimize the distance between $\tilde{\mathbf{C}}$ and $\mathbf{C}$. Whereas, it has been investigated~\cite{li2019single} that due to limited hardware support and influences of inevitable adverse shooting conditions, the given ground truth may also contain unpredictable biases, which misguide the derain tasks even the downstream tasks. The above phenomenon enlightens us to rethink the supervised clean data and refine them with the proposed auxiliary mirror attack. 

Specifically, inspired by the idea~\cite{sun2022rethinking} which establishes the correlation between restoration and objective detection tasks by generating pseudo ground truth for upstream restoration tasks, we here design an Auxiliary Mirror Attack  (AMA) generator denoted as $\mathcal{G}_{m}(\cdot|\boldsymbol{\delta})$ to generate the mirror attack of NAA aiming to minimize the attack objective $\mathcal{L}_{atk}$. In comparison with the negative impact of $\boldsymbol{\delta}_{n}$ generated by $\mathcal{G}_{n}(\cdot|\boldsymbol{\delta})$,  $\mathcal{G}_{m}(\cdot|\boldsymbol{\delta})$  is supposed to dynamically adjust the derain results with attack prior of the inner maximization objective, and add the “positive” perturbation to the clean image. Then the objective of PEARL framework with AMA can be further reformulated as:
\begin{equation}
	\begin{aligned}
		&\underset{\boldsymbol{\theta}}{\operatorname{min} } \   \mathcal{L}_{def}(\mathcal{F}(\mathbf{I} + \boldsymbol{\delta}_{n}|\boldsymbol{\theta}), \mathbf{C} + \boldsymbol{\delta}_{m} ) \\
		&s.t. \boldsymbol{\delta}\in \underset{\boldsymbol{\delta},\|\boldsymbol{\delta}\|_p \leq \epsilon}{\operatorname{argmax}}\  \mathcal{L}_{atk}(\mathcal{S}(\mathbf{I}+\boldsymbol{\delta}_{n}|\boldsymbol{\omega}),\mathbf{Y} ) ,
		\label{pirt_ama}
	\end{aligned}
\end{equation}
where $\boldsymbol{\delta}_{m}=\mathcal{G}_{m}(\mathcal{L}_{atk}(\mathcal{S}(\mathbf{I}+\boldsymbol{\delta}_{n}|\boldsymbol{\omega}),\mathbf{Y} )|\boldsymbol{\delta})$. We describe the whole pipeline of our PEARL framework with AMA in Fig.~\ref{pipeline}(b). In some degree, PEARL intends to train the derain model to decompose the clean image, thus defend the adversarial attack generated by $\mathcal{G}_{n}(\cdot|\boldsymbol{\delta})$, while AMA moves one more step to interpolate the mirror attack of $\mathcal{G}_{n}(\cdot|\boldsymbol{\delta})$ to the ground truth. Consequently, by minimizing $\mathcal{L}_{def}(\tilde{\mathbf{C}}, \mathbf{C}+\boldsymbol{\delta}_{m})$, our proposed framework with AMA turns the decomposition mapping in Eq.~\eqref{pirt_ama} to the following one:

\begin{equation}
	\begin{aligned}
	\mathbf{I}+\boldsymbol{\delta}_{n}\to \tilde{\mathbf{C}} +( \mathbf{R} +\boldsymbol{\delta}_{n}) \Rightarrow \ &\mathbf{I}+\boldsymbol{\delta}_{n}\to (\mathbf{C}+\boldsymbol{\delta}_{m}) +( \mathbf{R} +\boldsymbol{\delta}_{n})\\
	 \Rightarrow \ &\mathbf{I}+\boldsymbol{\delta}_{n}\to( {\mathbf{C}} +{\mathbf{R}}) +({\boldsymbol{\delta}_{n}}+{\boldsymbol{\delta}_{m}}).\\
\end{aligned}
	\label{dempose_map_amag}
\end{equation}
It can be observed that the generated $\boldsymbol{\delta}_{m}$ added in $\mathbf{C}$ serves as distribution prior of ${\boldsymbol{\delta}_{n}}$, which facilitates the robust learning of derain model in Eq.~\eqref{pirt_ama} based on its original decomposition function. Meanwhile, since AMA introduces the information of adversarial attack on segmentation model to the ground truth of derain model, the output results will consistently benefit the segmentation tasks to some extent.  

In comparison with the NAT framework in Fig.~\ref{pipeline} (a), which forms two cycles by alternatively optimizing the attack generator and segmentation model with $\mathcal{L}_{atk}$ and $\mathcal{L}_{def}$, our complete pipeline with both PEARL framework and AMA generator creates a new cycle by introducing AMA to the optimization of deraining model in Fig.~\ref{pipeline} (c). Besides, we also analyze the difference of training strategies between NAT frameworks and our PEARL with AMA to help understand how to update the attack and defense module in Fig.~\ref{pipeline} (d). In the next section, we will demonstrate the significant performance improvement and generalization performance of this new framework with different quantitative and qualitative metrics on derain and segmentation tasks.

\section{Experiments}
\subsection{Experimental Settings}

\begin{figure*}[htb]
	\centering
	\begin{tabular}{c}
		\includegraphics[height=11cm,width=17.13cm,trim=10 10 140 10,clip]{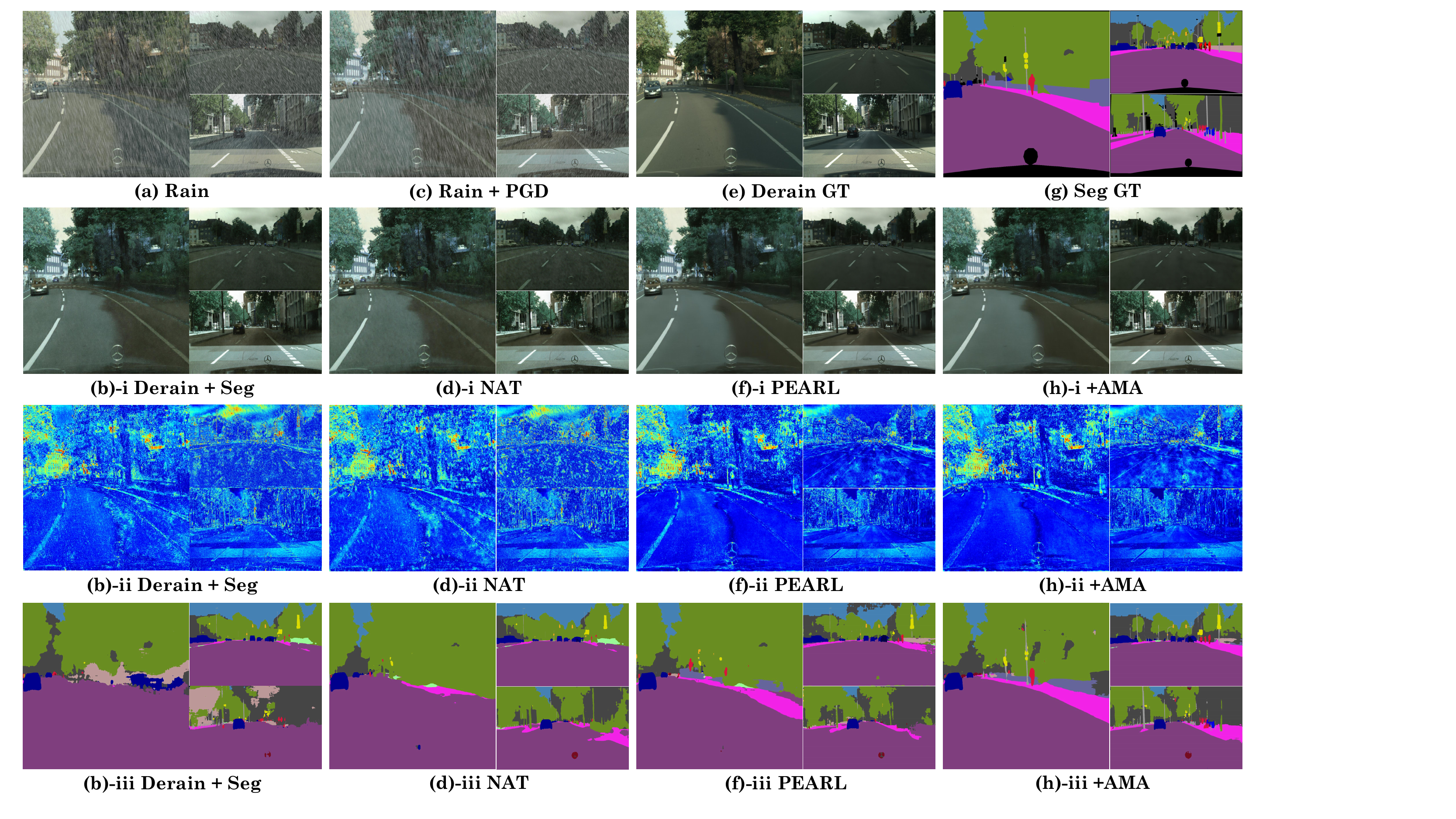} \\
	\end{tabular}
	\vspace{-0.3cm}
	\caption{Comparison of the deraining and segmentation results among different methods on synthesized Cityscapes dataset. The second to fourth rows of images represent the deraining results, heat map of the difference between the derained image and clean image, and the segmentation labels. }\label{fig_all_attack}
	\vspace{-0.2cm}
\end{figure*}

\paragraph{\textbf{Dataset and Model}} We implement our experiments based on two popular semantic segmentation datasets, including Cityscapes~\cite{cordts2016cityscapes} and PASCAL VOC 2012~\cite{everingham2010pascal}. In the following, we train the model based on the training dataset of Cityscapes, while both datasets are used for testing to verify the performance improvement and generalization ability of the proposed framework. Here we employ two widely used models, i.e.,   PSPNet~\cite{zhao2017pyramid} and DeepLabv3~\cite{chen2017rethinking} for the downstream segmentation task. ResNet50~\cite{he2016deep} is adopted as backbone feature extractor, and we follow the default setting of model configuration for training and testing. As for the derain models,  we implement four mainstream deraining models, TransWeather~\cite{valanarasu2022transweather}, MPRNet~\cite{zamir2021multi}, PReNet~\cite{ren2019progressive} and RESCAN~\cite{li2018recurrent} to verify the consistent performance of PEARL framework and its insensitivity to the architecture  of derain model.


\paragraph{\textbf{Degradation factors and Metrics}} For natural degradation factor (i.e. rain streaks), we manually synthesize rain streaks based on original Cityscapes and VOC dataset( the initial PSNR and SSIM are  17.45 / 0.5566). For artificially generated degradation factor (i.e. adversarial attack), we use BIM attack for training, while BIM, PGD, CW are used for testing the defense performance ($\epsilon=4/255$, $8/255$). As for the metrics, two type of pixel-wise Accuracy (overall accuracy allAcc and mean of class-wise accuracy mAcc) and mean of class-wise Intersection over Union (mIoU) are used to evaluate the performance of segmentation, which also reflects the robustness against different degradation factors. In addition, PSNR and SSIM are used for the low-level restoration tasks. More implementation details could be found in the supplementary materials.

%

\subsection{Experimental Results}


We first evaluate the performance of different strategies when both rain streaks and adversarial perturbation exists in the segmentation input. Generally speaking, we consider several basic strategies and our proposed framework to address this challenging task. We use Seg, Robust Seg and Derain + Seg to represent the basic model trained with clean image and two models for only handling rain streaks or adversarial examples. Meanwhile, we test the performance of our proposed NAT framework, PEARL framework and PEARL with AMA generator (denoted as +AMA). 

In Tab.~\ref{table_all_attack}, we consider BIM ($K=3,5,10$), PGD and C\&W attack constrained by $\ell_{\inf }$ norm together with the rain streaks to attack the segmentaion task on synthesized Cityscapes dataset. As it can be observed, both degradation factors could incur a sharp decline in the performance of downstream segmentation task. When the attack intensity is weak ($\epsilon=4/255$, the results can be found in the supplementary material), embedding the pretrained derain model may help protect the segmentation tasks to some extent. Whereas, once the attack intensity increases to $8/255$, the deraining model with little attack prior will also be affected by the perturbation, which causes serious performance decrease. Besides, as the adversarial robustness of the segmentation model improves, the perturbation generated by the same attack method also becomes stronger, which can be reflected in the decline of PSNR.

In contrast, the three proposed solutions, which take into account both factors, significantly promote the defense capability of segmentation tasks. Among these three solutions, PEARL framework (with AMA) gains much more improvement on both derain and segmentation metrics. Under a relatively weak attack ($\epsilon=4/255$), the effectiveness of AMA can not be clearly verified. As the intensity of adversarial attack increases ($\epsilon=8/255$), with a slight trade-off on deraining performance (0.1 decrease of PSNR), AMA enables better performance of downstream segmentation task on both mIoU and allAcc metrics. It is also worth noting that for unseen attacks (PGD and CW attack), PEARL framework together with AMA assistance still maintains a stable defense effect.


\begin{figure}[htt]
	\vspace{-0.2cm}
	\begin{center}
		\renewcommand\arraystretch{0.1}
		\begin{tabular}{c@{\extracolsep{0.1em}}c@{\extracolsep{0.1em}}}
			\multicolumn{2}{c}{\includegraphics[width=1.0\linewidth,trim=55 510 90 30,clip]{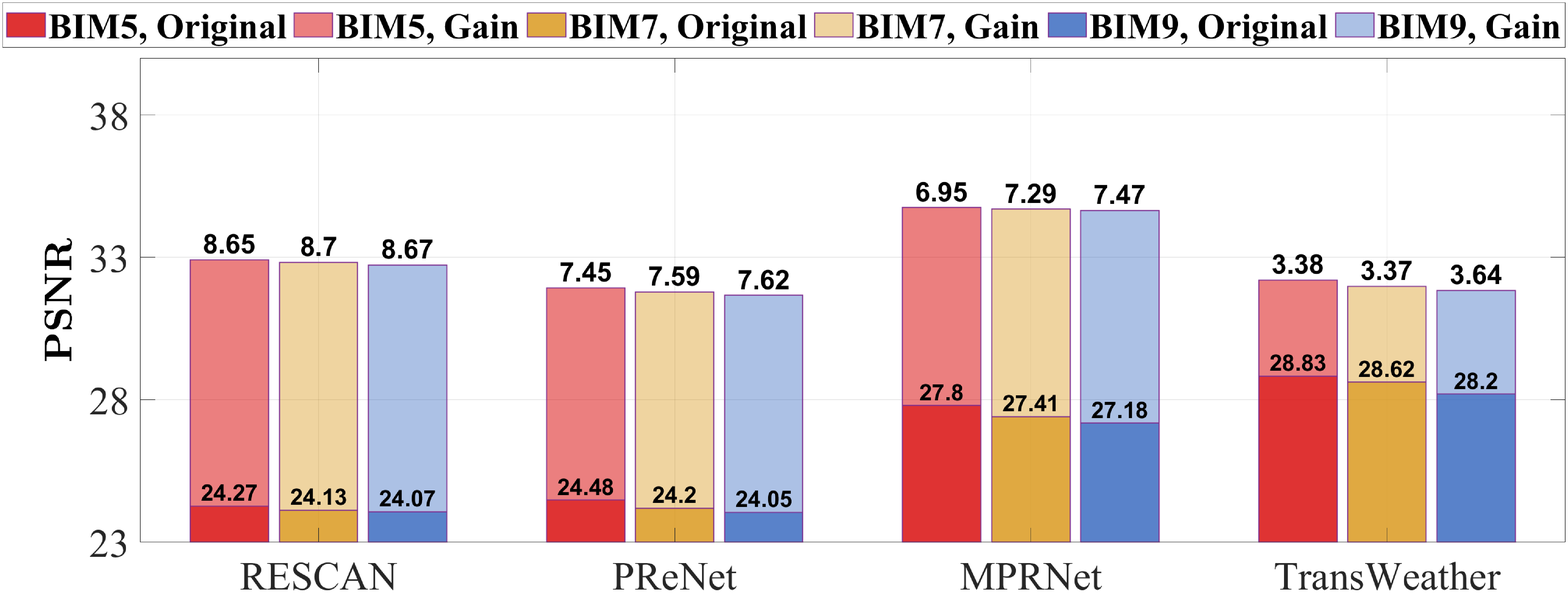}}\\
			\specialrule{0.0em}{1pt}{0pt}
			\includegraphics[height=0.095\textheight,width=0.5\linewidth,trim=66 0 70 44,clip]{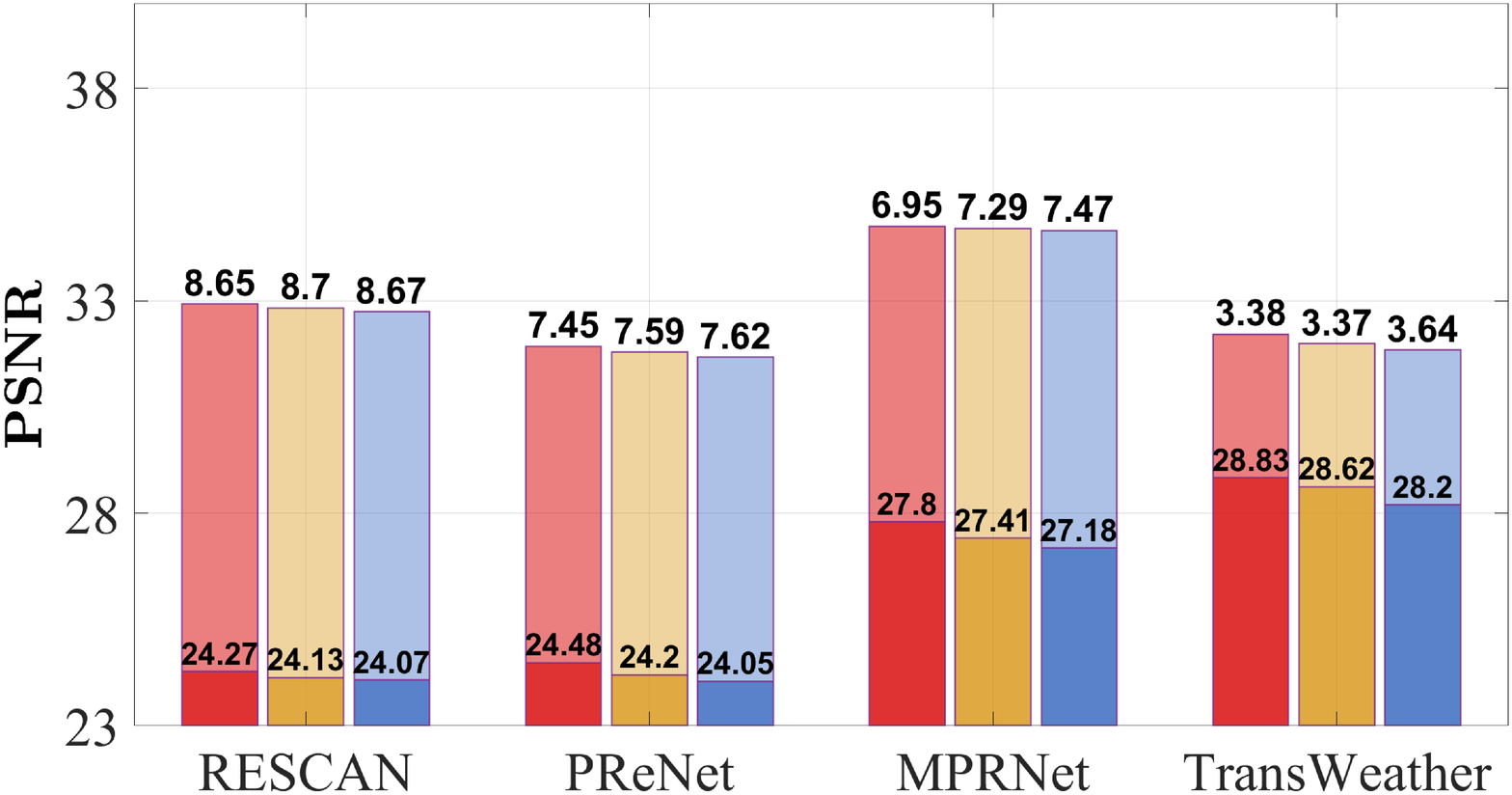}&
			\includegraphics[height=0.099\textheight,width=0.5\linewidth,trim=68 0 70 86,clip]{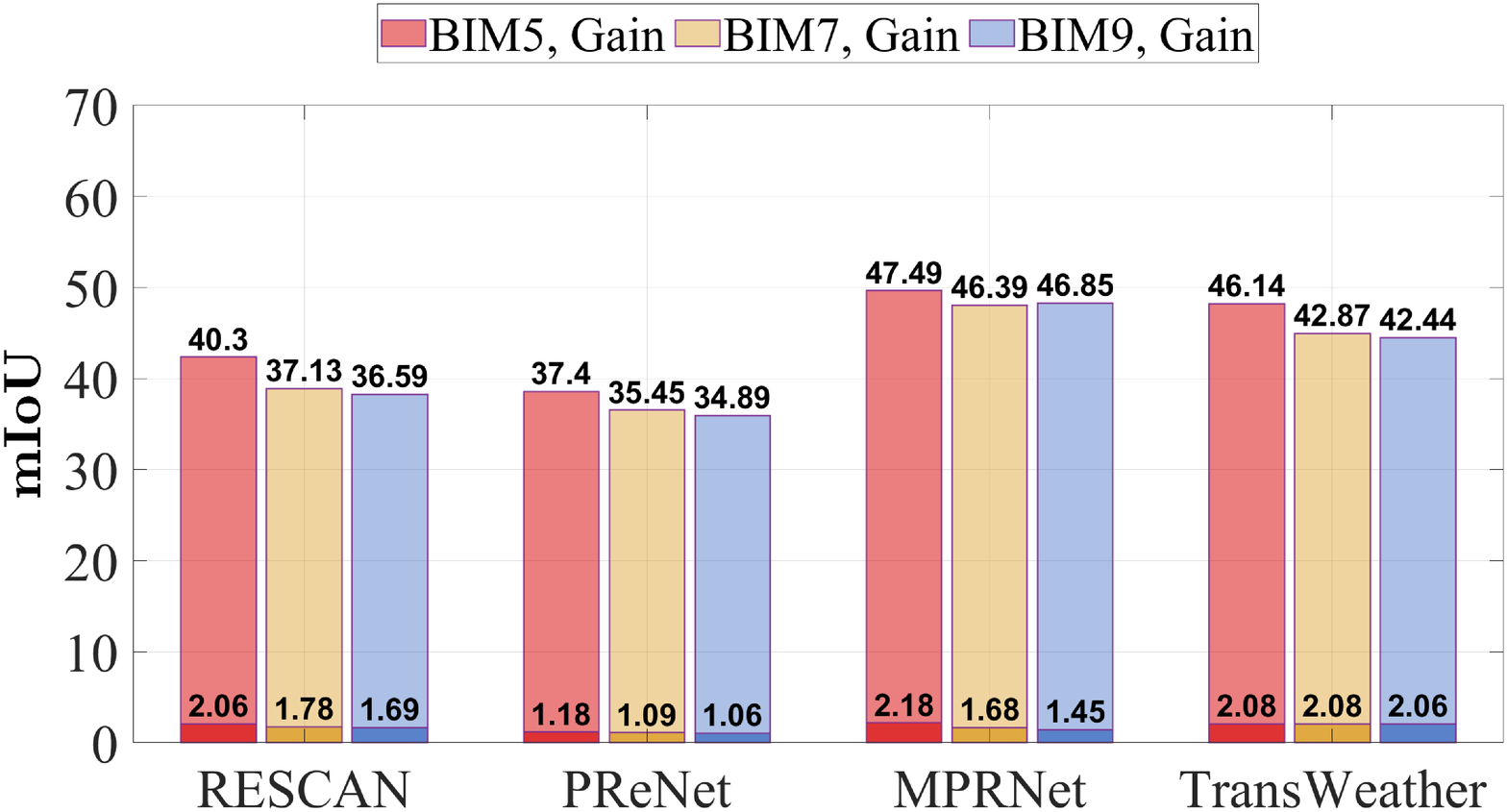}\\
		\end{tabular}
	\end{center}
	\vspace{-0.4cm}
	\caption{We illustrate the performance improvement of our PEARL Framework on PSNR and SSIM based on different attack intensities ($K=5,7,9$) and derain models, including RESCAN, PReNet, MPRNet and Transweather.}\label{fig:derain_methods}
	\vspace{-0.4cm}
\end{figure}

Furthermore, we also show the visualization results in Fig.~\ref{fig_all_attack} to demonstrate that our PEARL framework helps obtain higher quality derained images and effectively facilitates the downstream segmentation tasks to defend two degradation factors, which leads to better segmentation results. From the processed heat maps in the third row, it can be clearly seen that the output of deraining model trained by PEARL left much less noise than other solutions, which also demonstrates the effectiveness of PEARL framework to obtain derain images with better visual effects.

Then we adopt four state-of-the-art deraining methods to verify the insensitivity of PEARL framework to the architecture of derain models, and the results are shown in Fig.~\ref{fig:derain_methods}. We train these four models with the same strategy of PEARL with AMA in Fig.~\ref{pipeline} (d). It can be seen that our framework can not only improve the PSNR of these methods under different intensities of attack factors, but also significantly improve the downstream segmentation tasks to a large margin.

\begin{table}[htb]
	\renewcommand{\arraystretch}{1.2}
	\setlength{\tabcolsep}{0.6 mm}
	\caption{Reporting the defense performance of NAT, PEARL, and PEARL with AMA on the synthesized Cityscapes dataset.  }
	\label{table_example}
	\centering
	\begin{tabular}{c|cc|cc|cc|cc}
		\toprule[1.3pt]
		\toprule[0.9pt]
		\multirow{2}{*}{Methods} & \multicolumn{2}{c|}{Rain} & \multicolumn{2}{c|}{Rain + BIM} & \multicolumn{2}{c|}{Rain + PGD} & \multicolumn{2}{c}{Rain + CW} \\
		& mIoU        & PSNR        & mIoU           & PSNR           & mIoU           & PSNR           & mIoU          & PSNR          \\ \hline
		NAT                               & 51.94       & 31.38       & 33.58          & 28.03          & 43.92          & 30.15          & 30.89         & 27.59         \\
		PEARL                             & \textbf{58.39}       & 33.08       & 43.70          & 30.16          & 52.35          & \textbf{32.77}          & 38.02         & \textbf{31.90}         \\
		+AMA                         & 57.86       & \textbf{33.12}       & \textbf{52.51}          & \textbf{32.77}          & \textbf{53.68}          & 32.74          & \textbf{41.24}         & 31.76        
		\\ 
		\bottomrule[0.9pt]
		\bottomrule[1.3pt]
	\end{tabular}
\end{table}

\begin{figure}[htb]
	\begin{center}
		\renewcommand\arraystretch{0.8}
		\begin{tabular}{c}
			\includegraphics[height=0.25cm,width=8.5cm,trim=200 640 100 50,clip]{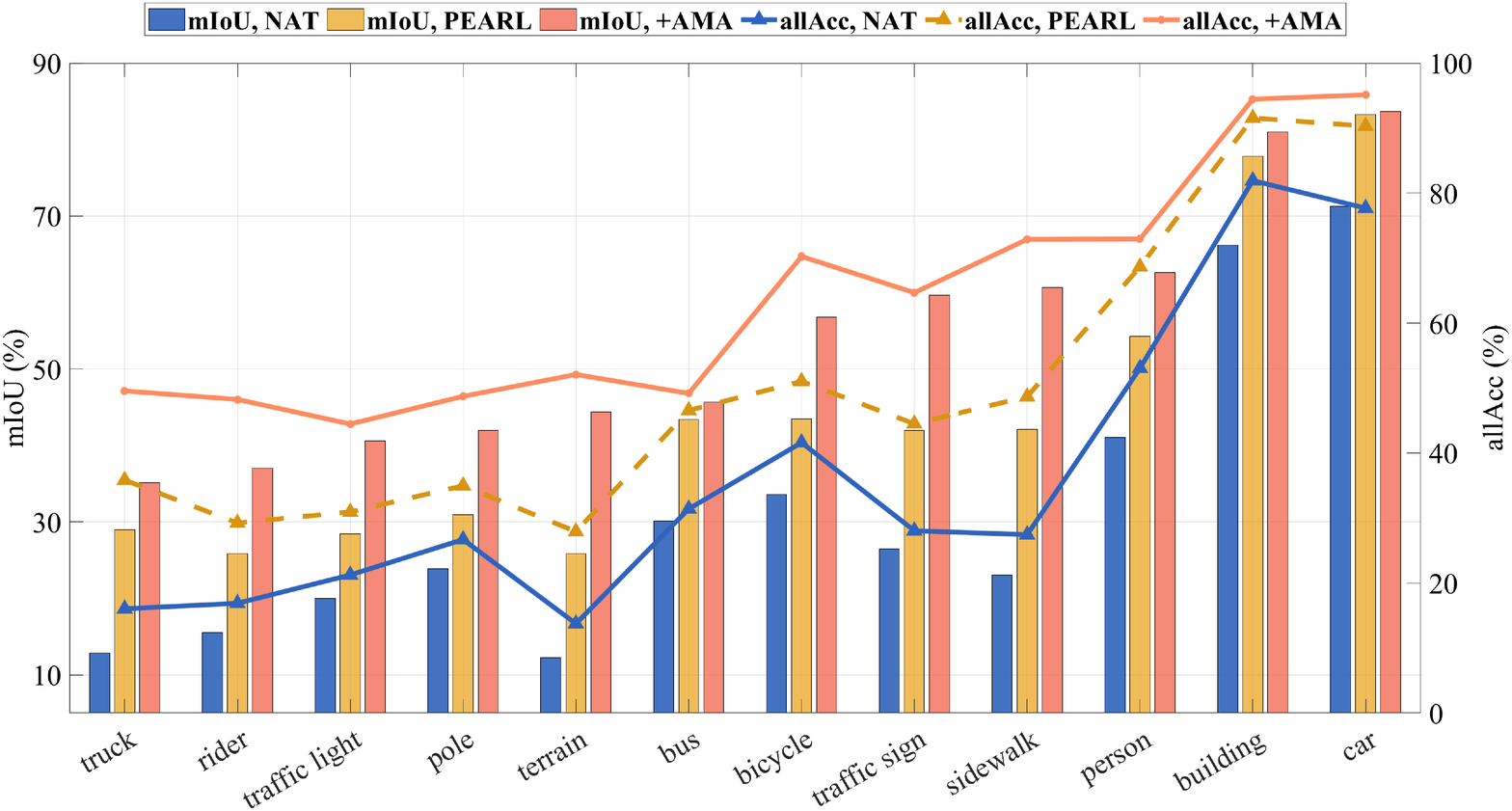}\\
		\end{tabular}
		\begin{tabular}{c}
			\includegraphics[height=3.7cm,width=8.5cm,trim=120 0 0. 90,clip]{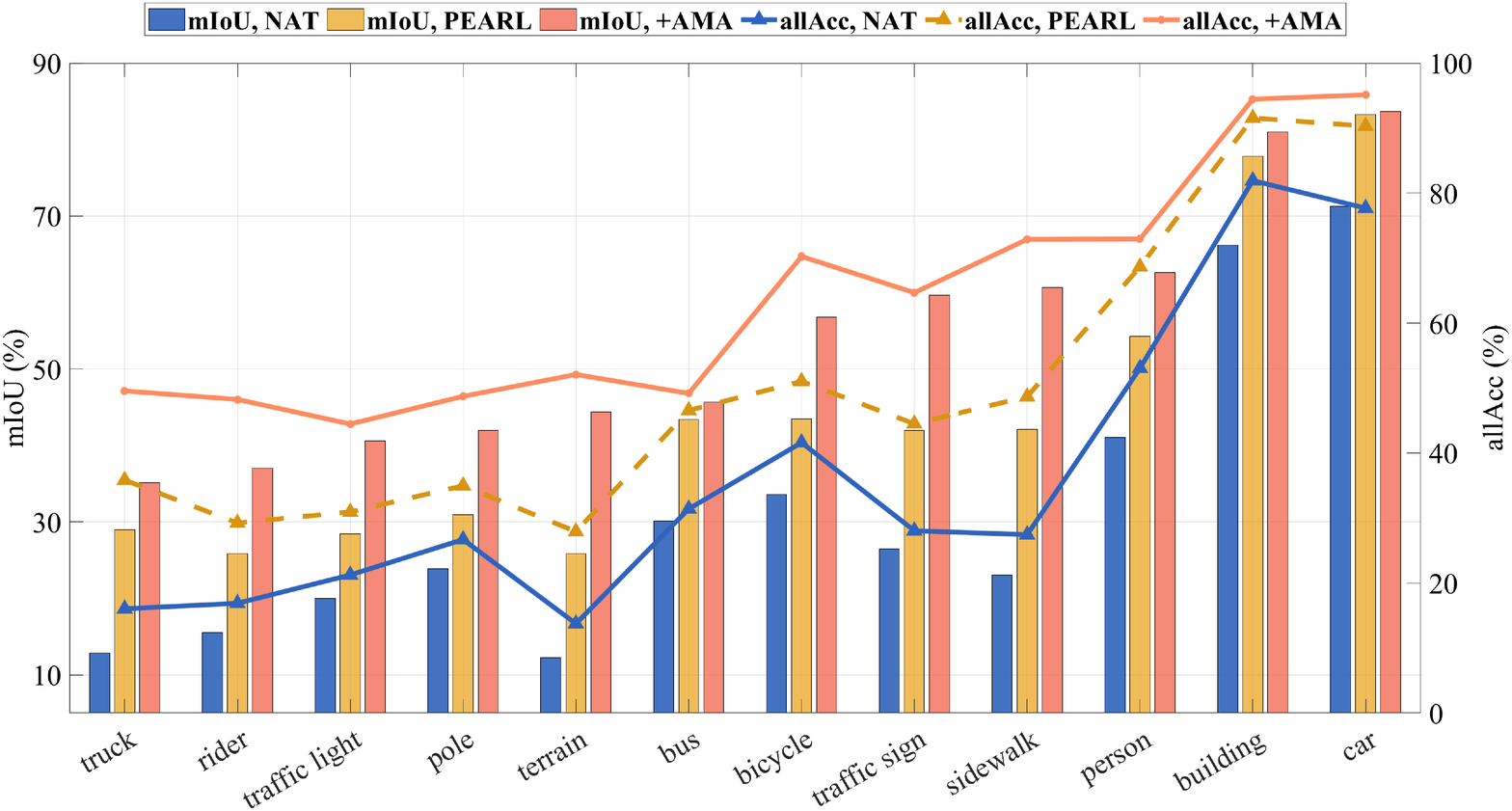}\\
		\end{tabular}
	\end{center}
	\vspace{-0.4cm}
	\caption{Illustrating the mIoU and allACC of different classes for NAT, PEARL and Pearl with AMA based on DeepLabv3.}\label{deeplab_iou_acc}
\end{figure}

Finally, we fix the trained deraining model and replace the PSPNet model with DeepLabv3 to show the generalization performance of derain model trained with PEARL to handle the adversarial attacks for the segmentation model. The results under different attacks and the segmentation results of different classes are shown in Table 2 and Figure 6 respectively. It can be seen that in the face of new downstream architecture, except for the unknown attack (CW), the deraining model trained by the PEARL framework and the AMA assistant can achieve a defense effect so close to the original PSPNet model.

\subsection{Ablation Study}
In essence, the motivation of PEARL framework together with AMA is to protect the downstream segmentation tasks from the impact of both natural degradation factor and artificially generated degradation factors. Here we conduct ablation experiments to analyze the practical effect of our framework to defend these degradation factor separately.

\begin{table}[htb]
	\renewcommand{\arraystretch}{1.2}
	\setlength{\tabcolsep}{1 mm}
	\caption{Results of different metrics with single degradation factor, i.e. rain streak.}
	\label{ablation_derain}
	\centering
	\begin{tabular}{c|lllll}
		\toprule[1.3pt]
		\toprule[0.9pt]
		Methods& mIoU  & mAcc  & allAcc & PSNR  & SSIM  \\ \hline
		Derain+Seg       & \textbf{37.52} & \textbf{39.64} & \textbf{97.00}  & 31.41 & 90.87 \\
		PEARL      & 31.96 & 36.37 & 96.01  & \textbf{33.13} & 92.69 \\
		+AMA & 31.79 & 36.10 & 96.53  & 33.06 & \textbf{93.05} \\ 
		\bottomrule[0.9pt]
		\bottomrule[1.3pt]
	\end{tabular}
\end{table}

 Specifically, we first validate the deraining model trained by our framework on images with only rain streaks in Tab.~\ref{ablation_derain}. It can be observed that the trade-off between accuracy on clean data and the robustness to defend adversarial attacks also influences the performance of derain model trained by PEARL and PEARL with AMA. When only the rain streaks exist in the input, the model trained by our framework also gains worse performance on these images without adversarial perturbation. But we are also surprised to find that the derain performance are further improved as extra bonus to obtain better visualization results.

 \begin{figure}[htb]
 	\begin{center}
 		\renewcommand\arraystretch{1.2}
 		\begin{tabular}{c}
 			\includegraphics[height=2.3cm,width=8.15cm,trim=10 0 0 0,clip]{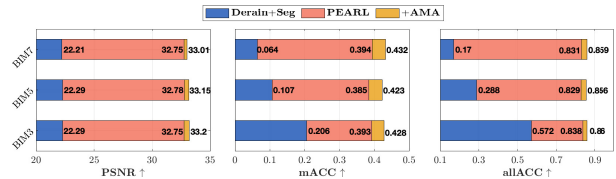}\\
 		\end{tabular}
 	\end{center}
 	\vspace{-0.4cm}
 	\caption{Illustrating the evaluation results of Derain + Seg, PEARL and PEARL with AMA under different attack intensities of BIM ($K=3, 5, 7$).}\label{ablation_attack}
 	\vspace{-0.2cm}
 \end{figure}
 
 As for Fig.~\ref{ablation_attack}, it comes to a conclusion that our framework indeed enables the deraining model to obtain the ability to eliminate adversarial perturbation under different attack intensity to a great extent, and AMA also further improved the segmentation results and quality of restored images when the rain no longer exists.

\subsection{Extension}

Last but not least, we also validate the generalization performance of the proposed framework across different datasets. Specifically, we transfer the deraining model in Table~\ref{table_all_attack} (trained on PSPNet and Cityscapes by PEARL framework) directly to PASCAL VOC dataset (the segmentation modeltrained on PASCAL VOC are also employed), and report the results in Table~\ref{table_transfer}. It can be seen that even in the face of unseen input data distribution and new downstream segmentation model, PEARL framework with the assistance of AMA, can still obtain significant performance in comparison with NAT.

\begin{table}[htb]
	\renewcommand{\arraystretch}{1.2}
	\setlength{\tabcolsep}{0.6 mm}
	\caption{Results of the defense performance on PASCAL VOC dataset. The derain model is the same as the one used in Table~\ref{table_all_attack}, while the segmentation model was replaced.}
	\label{table_transfer}
	\centering
	\begin{tabular}{c|cc|cc|cc|cc}
		\toprule[1.3pt]
		\toprule[0.9pt]
		\multirow{2}{*}{Methods} & \multicolumn{2}{c|}{Rain} & \multicolumn{2}{c|}{Rain + BIM} & \multicolumn{2}{c|}{Rain + PGD} & \multicolumn{2}{c}{Rain + CW} \\
		& mIoU        & Acc      & mIoU           & Acc         & mIoU           & Acc         & mIoU          & Acc        \\ \hline
		NAT                               & 53.05       & \textbf{79.71}      & 36.59          & 66.62          & 36.28          & 66.03          & 36.28         & 65.91         \\
		PEARL                             & \textbf{58.41}       & 72.84       & 43.16          & 73.50          & 43.32          & 73.89          & 43.41         & 73.86         \\
		+AMA                         & 58.38       & 72.89       & \textbf{43.54}          & \textbf{73.95}          & \textbf{43.88}          & \textbf{74.42}          & \textbf{43.87}         & \textbf{74.22}        \\
		\bottomrule[0.9pt]
		\bottomrule[1.3pt]
	\end{tabular}
 	\vspace{-0.4cm}
\end{table}

\section{Conclusion}

In this paper, we have addressed the robustness of semantic segmentation tasks in a general application scenario where the input image is affected by both natural degradation factors (i.e., rain streaks) and artificially generated degradation factors (i.e., adversarial attacks). Based on the unified understanding of the above degradation factors and analysis of proposed NAT framework, we introduced the PEARL framework, which leverages the adversarial robustness by transferring it to the derain model to simultaneously eliminate the influence of both rain streaks and adversarial perturbation. Moreover, we introduced the AMA generator to the PEARL framework, which provides positive information prior for the defense update as opposed to the NAA generator. We have shown the significant performance improvement of the PEARL framework for handling both types of degradation factors based on different derain and segmentation models. Furthermore, we have verified the generalization performance of the PEARL framework with AMA across different datasets.

\bibliographystyle{ACM-Reference-Format}
\bibliography{reference}

\appendix

%
%
%
%
%
%
%

\end{document}